\documentclass{article}

\usepackage[logo,copyright]{epic}

\usepackage[utf8]{inputenc} % allow utf-8 input
\usepackage[T1]{fontenc}    % use 8-bit T1 fonts
\usepackage{hyperref}       % hyperlinks
\usepackage{url}            % simple URL typesetting
\usepackage{booktabs}       % professional-quality tables
\usepackage{amsfonts}       % blackboard

\usepackage{nicefrac}       % compact symbols for 1/2, etc.
\usepackage{microtype}      % microtypography
\usepackage{fancyhdr}       % header
\usepackage{graphicx}       % graphics
\graphicspath{{media/}}     % organize your images and other figures under media/ folder
\usepackage{amsmath}
\usepackage[export]{adjustbox}
\usepackage{authblk} 
\usepackage{cite}
\usepackage{multirow}
\usepackage{xcolor}
\usepackage{tcolorbox}
\def\huggingface{\raisebox{-1.5pt}{\includegraphics[height=1.05em]{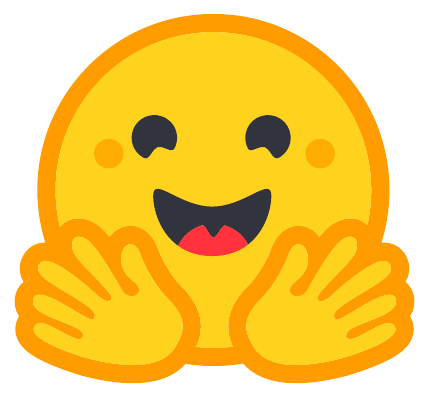}}}
\definecolor{LightGreen}{rgb}{0.0, 0.70, 0.0}
\def\github{\raisebox{-1.5pt}{\includegraphics[height=1.05em]{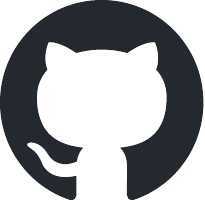}}}

\usepackage{multirow}
\usepackage{booktabs}
\newcommand{\hflink}{https://huggingface.co/maomaocun/dLLM-Var}
\newcommand{\githublink}{https://github.com/maomaocun/dLLM-Var}
\definecolor{mycolor}{RGB}{187, 64, 49}
\definecolor{linkcolor}{RGB}{187, 64, 49}

\usepackage{titlesec} % 您的模板已经加载，这里写出来以示清晰

\titleformat{\section}
  {\large\bfseries\headingfont\color{mycolor}}
  {\thesection.}
  {0.5em}
  {#1}
  []
\titleformat{name=\section,numberless}
  {\large\bfseries\headingfont\color{mycolor}}
  {}
  {0em}
  {#1}
  []
\titleformat{\subsection}
  {\bfseries\color{mycolor}}
  {\thesubsection.}
  {0.5em}
  {#1}
  []
\titleformat{\subsubsection}
  {\bfseries\itshape\color{mycolor}}
  {\thesubsubsection.}
  {0.5em}
  {#1}
  []
\titleformat{\paragraph}
  {\normalsize\bfseries\itshape\color{mycolor}}
  {\theparagraph}
  {1em}
  {#1}
\titleformat{\subparagraph}
  {\normalsize\bfseries\color{mycolor}}
  {\thesubparagraph}
  {1em}
  {#1}

\hypersetup{
    colorlinks=true,       % 确保链接是彩色的
    allcolors=linkcolor % 或者使用您定义的其他颜色
}
\title{Diffusion LLM with Native Variable Generation Lengths: Let [EOS] Lead the Way}

\author{
\vspace{5pt}
\begin{center}
  \large
  Yicun Yang$^1$ \textsuperscript{*} 
  \quad
  Cong Wang$^1$ 
  \quad
  Shaobo Wang$^1$ 
  \quad  
  Zichen Wen$^1$
  \quad
  Biqing Qi$^2$
  \quad
  Hanlin Xu$^3$
  \quad
  Linfeng Zhang$^1$ \textsuperscript{$\dagger$}   \\
  \vspace{5pt}
  \large
  Shanghai Jiao Tong University$^1$\quad Shanghai AI Lab$^2$\quad Huawei$^3$ \\
  
\end{center}
}

\begin{document}

\begingroup
\let\theabstract\relax
\let\absfont\relax
\maketitle
\endgroup

\footnotetext[1]{Project head: \texttt{\href{mailto:yangyicun187@gmail.com}{yangyicun187@gmail.com}}}
\footnotetext[2]{Corresponding author: \texttt{\href{mailto:zhanglinfeng@sjtu.edu.cn}{zhanglinfeng@sjtu.edu.cn}}}

\begin{tcolorbox}[
  colback=white,            % 背景颜色设置为纯白色
  colframe=Red!60!white,   % 边框颜色设置为柔和的浅蓝色
  arc=3mm,                  % 设置圆角，半径为3mm
  boxsep=3mm                % 边框与内容的距离
]
\vspace{-1cm}
\begin{center}
    \large\textbf{\textcolor{mycolor}{Abstract}}
\end{center}
% \vspace{-2mm} % 如果觉得标题和正文间距太大，可以取消这行注释来减少垂直间距
Diffusion-based large language models (dLLMs) have exhibited substantial potential for parallel text generation, which may enable more efficient generation compared to autoregressive models. However, current dLLMs 
suffer from \textbf{fixed generation lengths}, which indicates the generation lengths of dLLMs have to be determined before decoding as a hyper-parameter, leading to issues in efficiency and flexibility.
To solve these problems, in this work, we propose to train a diffusion LLM with native variable generation lengths, abbreviated as \textbf{dLLM-Var}. Concretely, we aim to train a model to accurately predict the [EOS] token in the generated text, which
makes a dLLM be able to natively infer in a block diffusion manner, while still maintaining the ability of global bi-directional (full) attention and high parallelism.
Experiments on standard benchmarks demonstrate that our method achieves a \textbf{30.1$\times$} speedup over traditional dLLM inference paradigms and a \textbf{2.4$\times$} speedup relative to autoregressive models such as Qwen and Llama.
Our method achieves higher accuracy and faster inference, elevating dLLMs beyond mere academic novelty and supporting their practical use in real-world applications. \emph{Codes and models have been released.
}
\vspace{10pt} % Adds some space before the contact info

% \begin{center}
    \small 
    \begin{tabular}{ll}
      \huggingface \ \texttt{Huggingface model:} & \url{\hflink}\\
      \github \ \texttt{Github repo:} & \url{\githublink}\\
    \end{tabular}
% \end{center}
\end{tcolorbox}

\begin{figure}[htbp]
    \centering
    \includegraphics[width=0.99\linewidth]{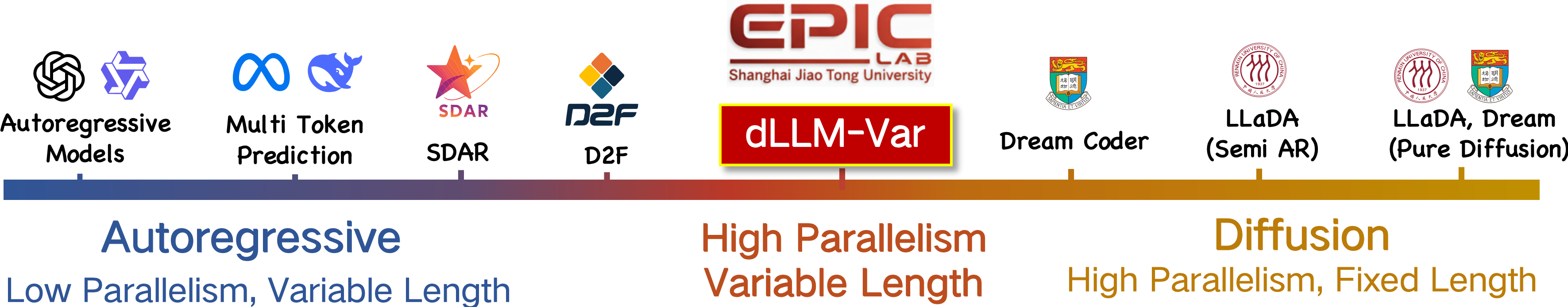}  
    \caption{\textbf{Overview of Probabilistic Modeling Paradigms for Text Generation}: Evolution from Autoregressive to Diffusion-Based Approaches. \textbf{AR \& MTP}       (left): Low parallelism,  variable generation. \textbf{Vanilla dLLMs} (right): High parallelism, fixed lengths.\textbf{ dLLM-Var} (middle): variable generation lengths while maintaining parallelism.}
    \label{fig:compare_method}
\end{figure}

\section{Introduction}
Recently, a multitude of diffusion-based large language models (dLLMs) have emerged~\cite{nie2025LLaDA,you2025llada,zhu2025llada,song2025seed,dream2025,xie2025dream,gong2025diffucoder,JetAstra2025,wu2025fastdllmv2efficientblockdiffusion}, positioning themselves as strong competitors to GPT-like models in the realm of scaling up. Concurrently, the pursuit of efficient and straightforward inference mechanisms for dLLMs represents a highly promising research avenue. Compared with the multi-token prediction~\cite{gloeckle2024better} and speculative decoding in autoregressive models, dLLMs have demonstrated the native ability of parallel decoding without any additional modules.
Recent works including SDAR~\cite{wu2025fastdllmtrainingfreeaccelerationdiffusion}, Fast-dLLM v2~\cite{wu2025fastdllmv2efficientblockdiffusion}, and D2F~\cite{wang2025diffusion} have shown significantly higher decoding speeds than AR models. However, success comes with still great challenges:

\noindent\textbf{Diffusion LLMs suffer from Fixed Length:}  dLLMs suffer from fixed generation lengths, which indicates that the generation length of them must be pre-defined before decoding as a hyper-parameter, which leads to abundant problems. For instance, an overlong generation length usually leads to repetition and hallucination, as well as abundant computation costs. On the other hand, an overly short generation length tends to have a negative influence on model accuracy, especially in reasoning tasks. Besides, in some applications, such as OCR, it is impossible to estimate the generation lengths before computation. As a result, the fixed generation lengths have limited the practical application of dLLMs.

\noindent\textbf{Block Diffusion Models Suffer from Lower Parallelism.} To solve this problem, several recent works have introduced the block diffusion models, where tokens in the same block are parallel decoded while blocks are organized in an autoregressive manner. Block diffusion models achieve the ability of any-length generation by inheriting the block-wise autoregressive generation. 
However, their parallel potential is severely constrained by their block sizes, which must be decided during training. Besides, as discussed by SDAR, it can be difficult to train a block diffusion with a very large block size, which further limits its upper bounds of parallelism. Besides, since the tokens in the following blocks do not influence tokens in previous works, it is not possible for block diffusion models to perform self-correction, which is a valuable ability for dLLMs.

%Although some dLLMs~\cite{nie2025LLaDA,dream2025} have significant potential for parallel generation, they cannot achieve satisfactory arbitrary-length generation, hindering their practical deployment in industry. As illustrated in Figure~\ref{fig:compare_method} (left), these autoregressive and multi-token prediction approaches suffer from low parallelism due to sequential generation.

\begin{figure}[tb!]
    \centering
    \includegraphics[width=1.0\linewidth]{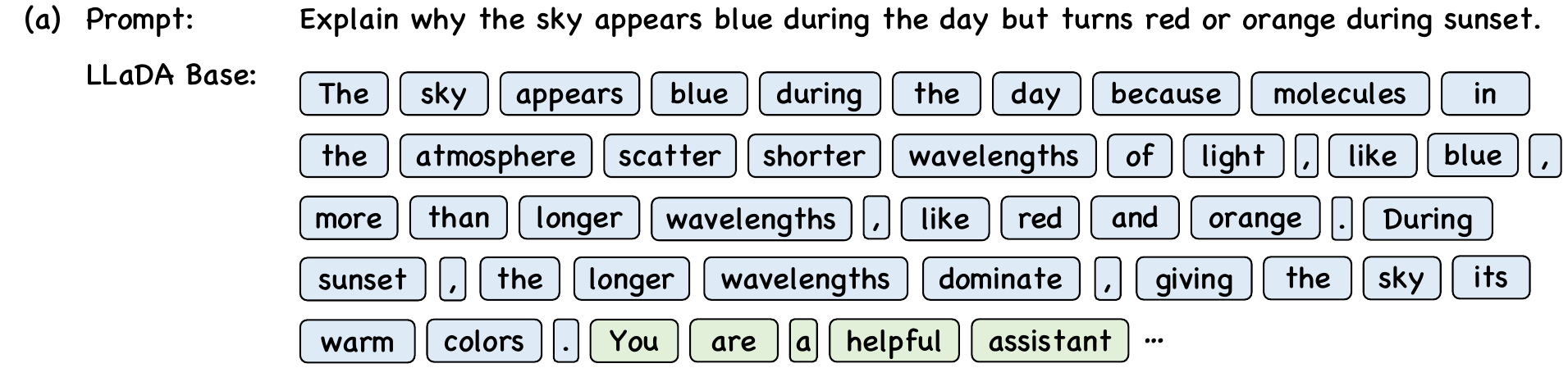}
    \includegraphics[width=0.99\linewidth]{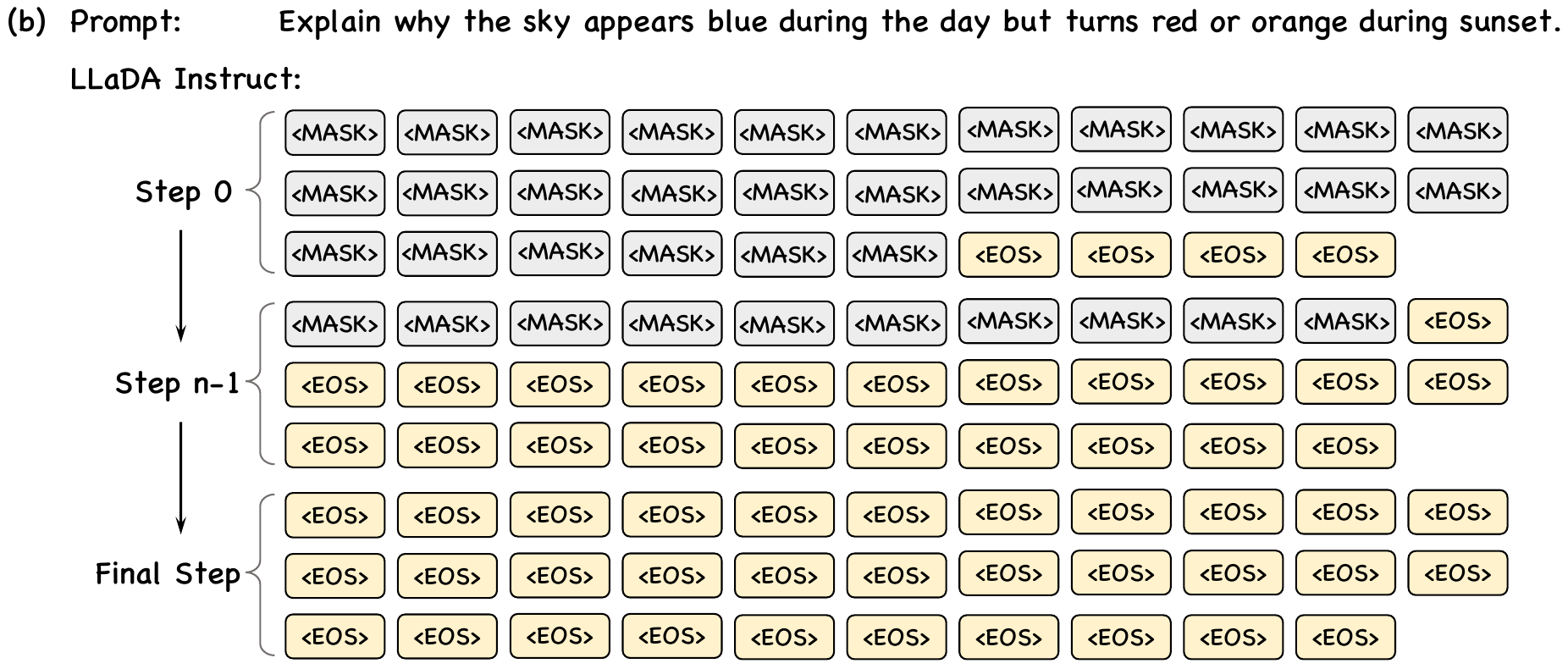}
    \caption{(a) LLaDA Base: generates irrelevant content without timely EOS token, complicating response extraction. (b) LLaDA Instruct under pure diffusion: fails to produce any effective text, prematurely filling trailing masks with EOS tokens.}
    \label{fig:llada_base_bad_case}
\end{figure}

The previous problems raise the question: \emph{is it possible to train a dLLM with variable generation length while still maintaining great parallelism?} Ideally, this can be achieved by inferring a dLLM in a block diffusion manner, which indicates continuously padding a block of <mask> tokens until the generation of the [EOS] token (\emph{i.e.}, the special token for termination), while still maintaining the bidirectional attention of dLLM. To achieve this, the model should be capable of accurately identifying the [EOS] token during decoding. In this paper, we first dive into the failure of [EOS] generation for the base model and the instruction model of LLaDA. As shown in Figure~\ref{fig:llada_base_bad_case}, the base model usually can not generate the [EOS] token even if it has completed its answer. Instead, it usually keeps generating grammatically correct tokens but makes no sense. For the model after SFT (\emph{i.e.,} the instruction model), without special limitation on the decoding tokens, the model usually tends to first decode all tokens in the tail to [EOS] tokens and then gradually generate [EOS] tokens at the beginning positions, losing the ability to generate any meaningful words. These phenomena demonstrate that current dLLMs do not have the ability for precise [EOS] prediction, which further explains their limitation on fixed-length generation.

 To address this problem, this paper proposes to train a dLLM with native ability of variable-length generation (dLLM-Var) by making it accurately decode the [EOS] token. Building on this insight, we introduce a novel training paradigm that empowers dLLMs to replicate the generative efficacy of block diffusion~\cite{arriola2025block} under full-attention regimes, while harnessing key-value (KV) caching for inference.
 dLLM-Var consists of two strategies. Firstly, we employ a deterministic noise scheduling for [EOS] tokens, where the EOS token is consistently replaced with a [MASK] token during training.
 Secondly, we propose to package multiple samples into a single sequence and train the dLLM over them without special attention masks, which enables the dLLM to understand the contextual function of the termination token, making it be able to predict [EOS] token at the proper position \emph{i.e., } the position that a context has been ended while a unrelated context will begin. 
 
In summary, this paper has the following contributions.
\begin{itemize}[leftmargin=*, topsep=3pt, itemsep=2pt]
\item We propose dLLM-Var, which introduce two SFT training strategies for dLLM, making a pretrained dLLM able to be inferred in a block diffusion manner, achieving native variable-length generation. Compared with the traditional block diffusion model, it maintains the bi-directional attention, making it possible for more flexible applications such as editing.
\item Based on the block diffusion-style generation, dLLM becomes more compatible with KV cache, avoiding the requirements for complex KV cache design, such as KV refreshing, delaying, and recomputing in previous works, leading to better efficiency and more elegant implementation.
\item Extensive experiment results demonstrate the performance of dLLM-Var, achieving comparable and even better performance than a dLLM with optimal fixed-generation length with over 30.1$\times$ acceleration, 2.4$\times$ faster than AR models. Its efficiency can be further improved based on step distillation\cite{chen2025dparallel}.
\end{itemize}
\section{Related Work}

\subsection{Diffusion-based Large Language Models}
Diffusion-based large language models (dLLMs) have recently gained prominence as scalable alternatives to autoregressive transformers, leveraging iterative denoising processes for text generation~\cite{nie2025LLaDA,you2025llada,zhu2025llada,song2025seed,dream2025,xie2025dream}. These models offer advantages in parallelism but face challenges in handling variable-length sequences and efficient caching mechanisms. Early works focused on adapting diffusion principles from continuous domains to discrete text spaces, enabling non-autoregressive generation with reduced latency compared to GPT-like architectures.

\subsection{KV Cache Management in dLLMs}
Key-value (KV) cache optimization is critical for efficient inference in large language models, and dLLMs introduce unique challenges due to their iterative sampling nature. Techniques such as dLLM-Cache~\cite{liu2025dllm} explore KV cache refreshing to mitigate redundancy in diffusion steps, while Fast-dLLM~\cite{wu2025fastdllmtrainingfreeaccelerationdiffusion} introduces training-free acceleration via block-wise updates. Similarly, dKV-Cache~\cite{jiang2025d2cache} proposes dual caching strategies to balance memory usage and generation quality. These methods enhance KV cache utilization in dLLMs~\cite{wei2025accelerating,dinfer}, yet they often require custom operators and struggle with fixed-length constraints.
\begin{figure}[tb!]
    \centering
    \includegraphics[width=1.0\linewidth]{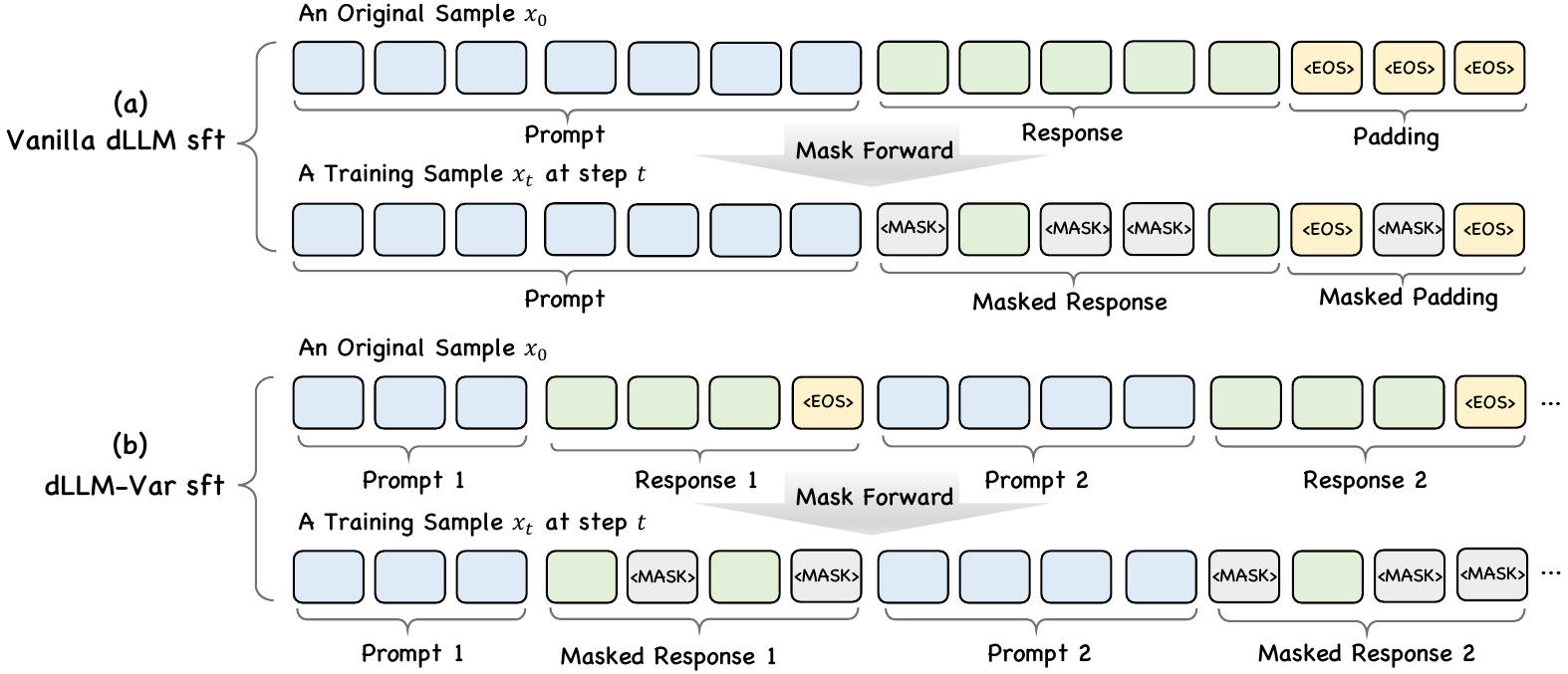}
            \caption{During the masking forward process of dLLM-Var, tokens in the prompt are never masked. In the response section, tokens are replaced with a <mask> token based on a probability, while the final EOS token is always masked.}
    \label{fig:method}
\end{figure}

\subsection{Flexible-Length Generation}
Facilitating variable-length outputs is crucial for the practical deployment of generative models. Methods such as DreamOn~\cite{Dreamon2025}, DAEDAL~\cite{li2025beyond}, and FlexMDMs~\cite{kim2025any} employ specialized tokens to dynamically elongate sequences. However, these techniques introduce overheads arising from diminished effective token density and protracted inference times. Such token-centric extensions underscore the inherent trade-offs between preserving diffusion stability and enabling unbounded generation.

\subsection{Block Diffusion and Attention Mechanisms}
Block diffusion inference paradigms, such as Block Diffusion~\cite{arriola2025block} and SDAR~\cite{JetAstra2025}, facilitate KV cache reuse and flexible-length generation. However, their reliance on specialized attention masks imposes significant drawbacks, including a twofold increase in computational training costs and restrictive block sizes (e.g., 4-8 tokens). Our work circumvents these limitations by enabling the effects of block diffusion under a full attention mechanism, thus promoting cache efficiency without requiring specialized masks. Furthermore, this architectural choice is crucial for future advancements. Recent works on self-reflective remasking~\cite{huang2025don,wang2025remasking} highlight the potential for self-correction by modifying previously generated tokens. This capability is fundamentally incompatible with block attention masks, which prevent revisiting prior blocks. In contrast, full attention provides the unrestricted access necessary for such corrective procedures. Consequently, full attention emerges not merely as an alternative, but as an essential requirement for this promising class of self-correcting dLLMs.

\section{Method}
In this section, we detail the core components of dLLM-Var, our proposed training framework for dLLMs. Unlike other dLLMs training methods~\cite{zhu2025llada,dream2025}, dLLM-Var
introduces two key innovations to enable flexible length generation and efficient KV cache reuse: (1) a fixed masking schedule for termination tokens to promote contextual awareness of sequence boundaries, and (2) multi-sample packing with full attention to enhance the model's understanding of termination signals across unrelated contexts. These modifications enable dLLMs' block diffusion during inference to truly understand whether the output text needs to terminate, rather than simply adding the termination tokens at the end of the first block.

\subsection{Fixed Masking for Termination Tokens}
During training, for a sequence \( x_0 = (x_0[0], \dots, x_0[L-1]) \), 
a noise level \( t \) is sampled uniformly from \( [0, 1] \), and each token \( x_0[i] \) 
is independently replaced with [MASK] with probability \( t \), yielding a noisy sequence \( x_t \). 
The model is then trained to predict \( x_t \) back to \( x_0 \) conditioned on the prompt.

During inference, termination tokens (e.g., EOS) must be generated by the model at appropriate positions 
and are not part of the input condition. To encourage this, we modify the masking schedule such that 
EOS tokens are always masked during training. Formally, the masking probability for position \( i \) is:

\[
p\bigl(x_t[i]=[\mathrm{MASK}]\mid x_0[i]\bigr) =
\left\{
\begin{array}{ll}
t, & \text{if } x_0[i] \neq \mathrm{EOS},\\[4pt]
1, & \text{if } x_0[i] = \mathrm{EOS}.
\end{array}
\right.
\]

Here, $x_0$ denotes the original (clean) sample, $x_t$ denotes the noisy version corresponding to the noise level $t$, and $i \in \{0, \dots, L-1\}$ indexes the token position within the sequence. This ensures that the model learns to output the termination token within a single conversation.

\begin{figure}[tb!]
    \centering
    \includegraphics[width=1.0\linewidth]{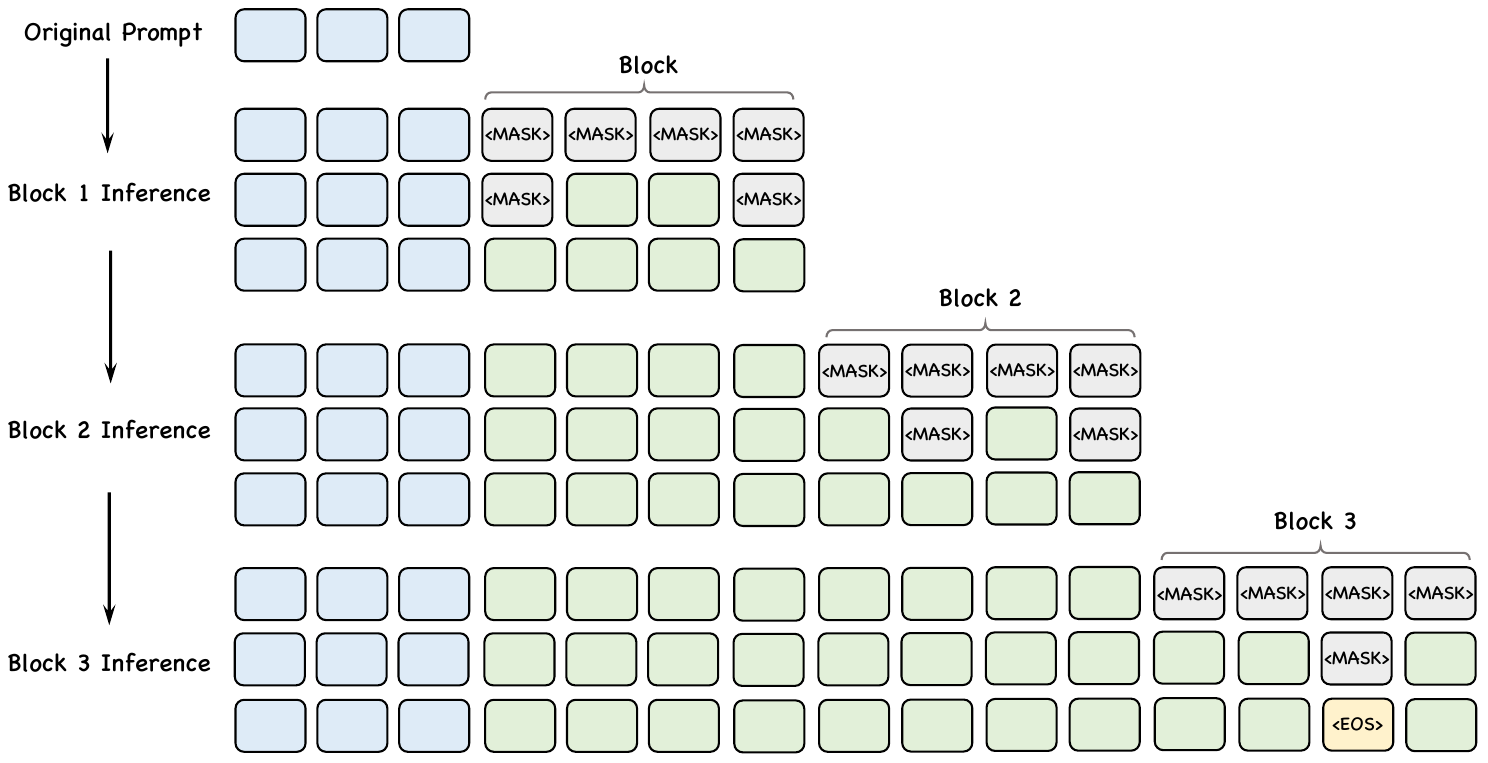}
    \vspace{-15pt}
    \caption{The inference process of dLLM-Var. For the prompt and the already generated blocks, they will be stored in the form of a KV cache to accelerate the model’s inference.}
    \label{fig:generate_process}
\end{figure}
\subsection{Multi-Sample Packing with Full Attention}
Merely fixing the masking for EOS tokens is insufficient, as it may lead the model to naively append EOS at the end of the first block without understanding contextual semantics. To address this, we introduce multi-sample packing: unlike standard single-sample or multi-turn supervised fine-tuning (SFT) in dLLMs \cite{nie2025LLaDA,dream2025}, we randomly sample \( N \) dialogue pairs \( \{(p^{(k)}, r^{(k)})\}_{k=1}^N \) from the dataset and concatenate them into a single training sequence of length \( L \) , separated by EOS tokens.

For the prompt parts \( p^{(k)} \), which serve as generation conditions, we set the masking probability to 0. For the response parts \( r^{(k)} \), masking follows the  \( t \)-based schedule, and all separating EOS tokens are always masked. Crucially, we apply full attention across the entire concatenated sequence, without any attention masks.

This packing exposes the model to multiple unrelated contexts separated solely by EOS, forcing it to learn the semantic role of EOS for generations, thereby enabling it to correctly output termination token. Surprisingly, this concatenation of unrelated samples does not lead to any degradation in model performance.

\subsection{Inference with Block Diffusion and KV Cache Reuse}
During inference, we adopt a block-wise diffusion process akin to Block diffusion~\cite{arriola2025block}, but under full attention without specialized masks. Given a prompt, we progressively append blocks initialized entirely with [MASK] tokens. The next block is only added after the current block is fully unmasked and contains no EOS token, where the block size is arbitrary. Through the above method, we can achieve generation of arbitrary lengths. See Figure~\ref{fig:generate_process} for an illustration of the inference process.

Analogous to autoregressive models, we consider that the prompt and the already generated parts have fixed semantic information under the premise that they do not need to be modified. Thus, reusing their KV caches does not compromise generation quality while substantially reducing redundant computations. Specifically, for the prompt and fully unmasked response blocks, we directly reuse their  KV caches to accelerate inference.
To enhance parallelism, we set a default block size of 64 tokens. Additionally, we incorporate a confidence threshold \cite{wu2025fastdllmtrainingfreeaccelerationdiffusion} of 0.9 for parallel decoding decisions, balancing between inference speed and output quality.

\begin{figure}[tb!]
    \centering
    \includegraphics[width=1.0\linewidth, height=0.4\linewidth]{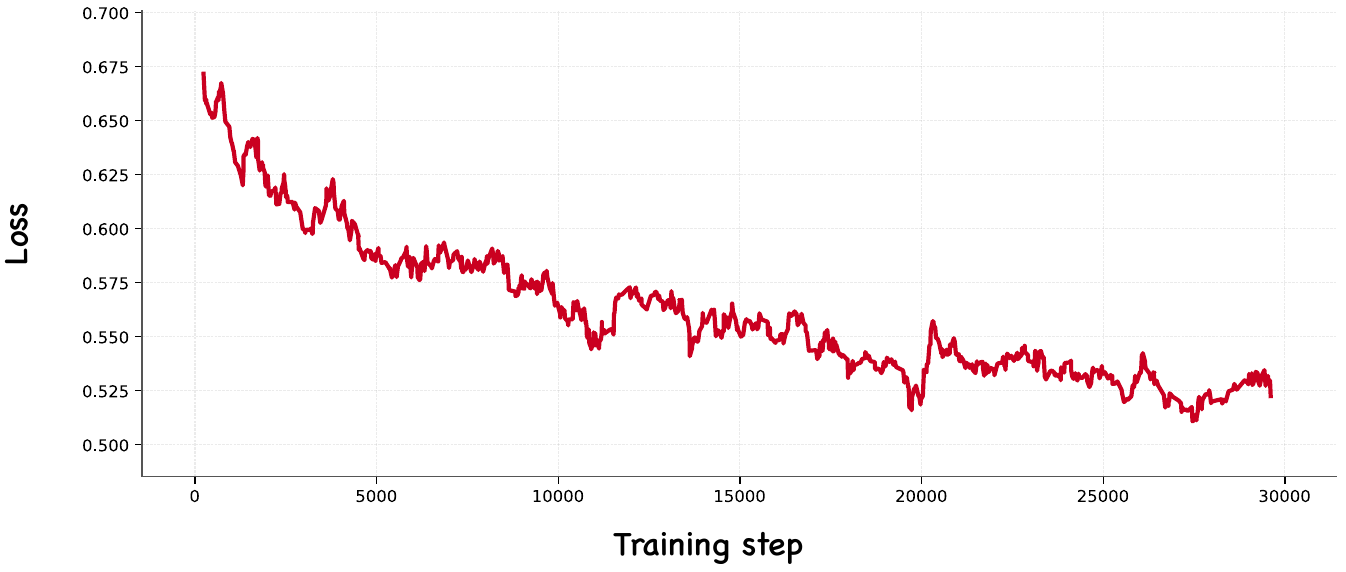}
    \vspace{-10pt}
    \captionof{figure}{\centering Training Loss curve during our training of dLLM-Var.}
    \label{fig:loss}
\end{figure}

\section{Experiments}

\subsection{Training Setting}
We trained the dLLM-Var model using 6.4 million supervised fine-tuning (SFT) dataset pairs over 3 epochs on the DeepSpeed ZeRO-2 framework across 8 nodes with 64 GPUs. The global batch size was 1 million tokens, with each sample having a length of 8192 tokens. The learning rate was set to \(1 \times 10^{-5}\), with a cosine warm-up schedule over the first 50 steps and no subsequent decay. To accelerate training, we employed FP8 mixed precision via Transformers Engines, during which no loss spikes were observed.

\subsection{Evaluation Setting}
During the evaluation experiments, the maximum generation length was set to 1024 for all setups, and the speedup ratio was obtained by comparing the time required to complete the model generation distribution for the evaluation tasks. All inference experiments were conducted on a single GPU, using bf16 precision, and without any additional optimizations to the PyTorch code.

\begin{figure}[tb!]
    \centering
    \includegraphics[width=0.99\linewidth]{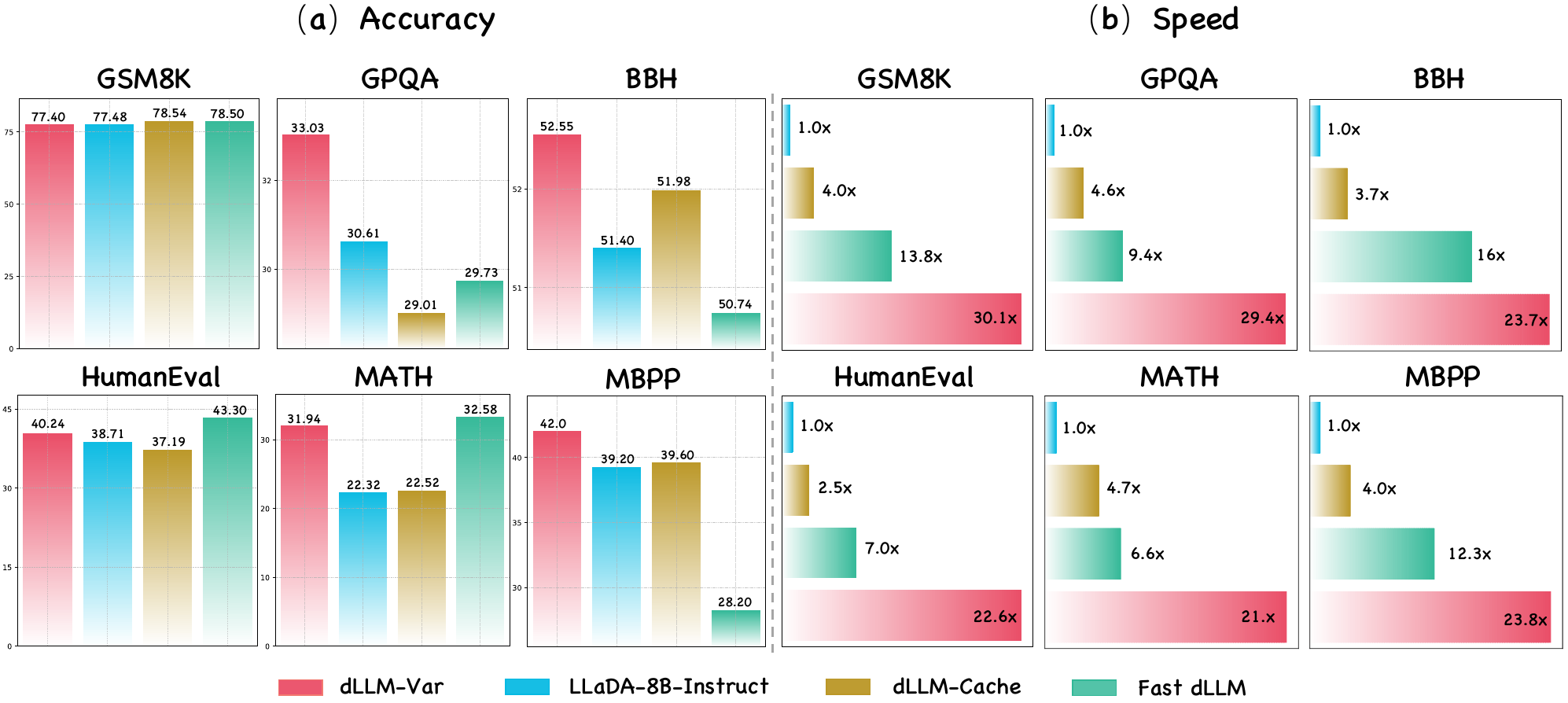}
    \caption{\textbf{Performance comparison of dLLM-Var against baseline methods.} (a) Accuracy comparison on six standard benchmarks. The results show that dLLM-Var maintains competitive accuracy while significantly improving inference speed. (b) Speed-up ratio relative to LLaDA-8B-Instruct. dLLM-Var demonstrates substantial acceleration, achieving up to a 30.1x speed-up. }
    \label{fig:acc_speed}
\end{figure}

\begin{table}[tb!]
\centering
\caption{Accuracy and Inference Speed Across Diverse Benchmarks. We tested several models with their corresponding inference methods: Semi-ar diffusion (for LLaDA-8B-Instruct, dLLM-cache, and fast-dLLM), Pure diffusion (for LLaDA-8B-Base and dLLM-Var-pd), and our proposed Block diffusion (for dLLM-Var-bd). Speed tests were performed with a maximum generation length of 1024 using official baseline scripts. The ``k-shot'' designation in the benchmarks specifies the number of few-shot examples per task.}
\label{tab:benchmark_comparison_revised}
\renewcommand{\arraystretch}{1.4} % Row height
\setlength{\tabcolsep}{4pt}      % Column spacing
\small                           % Font size
\resizebox{\textwidth}{!}{
\begin{tabular}{c|c|cc|cc|cc|cc|cc|cc}
\toprule
\multirow{2}{*}{\textbf{Model}} & \multirow{2}{*}{\textbf{\shortstack{Inference \\ Method}}} & \multicolumn{2}{c|}{\textbf{GSM8k (5-shot)}} & \multicolumn{2}{c|}{\textbf{GPQA (5-shot)}} & \multicolumn{2}{c|}{\textbf{BBH (3-shot)}} & \multicolumn{2}{c|}{\textbf{MATH (3-shot)}} & \multicolumn{2}{c|}{\textbf{HumanEval (0-shot)}} & \multicolumn{2}{c}{\textbf{MBPP (3-shot)}} \\ 
\cmidrule(lr){3-4} \cmidrule(lr){5-6} \cmidrule(lr){7-8} \cmidrule(lr){9-10} \cmidrule(lr){11-12} \cmidrule(lr){13-14}
& & \textbf{Acc.} & \textbf{Speed} & \textbf{Acc.} & \textbf{Speed} & \textbf{Acc.} & \textbf{Speed} & \textbf{Acc.} & \textbf{Speed} & \textbf{Acc.} & \textbf{Speed} & \textbf{Acc.} & \textbf{Speed} \\ 

\midrule
Qwen3-8B & Autoregressive & \textcolor{gray}{89.84} & 17.8$\times$ & \textcolor{gray}{44.44} & 18.7$\times$ & \textcolor{gray}{78.40} & 12.3$\times$ & \textcolor{gray}{60.80} &15.3$\times$ & \textcolor{gray}{65.93} & 9.3$\times$ & \textcolor{gray}{69.80} &11.3$\times$ \\ 
\midrule
Llama-8B-Instruct & Autoregressive & \textcolor{gray}{78.30} & 18.7$\times$ & \textcolor{gray}{31.90} & 19.2$\times$ & \textcolor{gray}{61.10} & 13.6$\times$ & \textcolor{gray}{29.60} & 18.5$\times$ & \textcolor{gray}{59.80} & 10.4$\times$ & \textcolor{gray}{57.60} & 12.3$\times$ \\ 
\midrule
\midrule
LLaDA-8B-Base & Pure diffusion & 69.06 & 1.0$\times$ & 31.91 & 1.0$\times$ & 44.77 & 1.0$\times$ & 30.84 & 1.0$\times$ & 32.92 & 1.0$\times$ & 40.80 & 1.0$\times$ \\ 
\midrule
LLaDA-8B-Instruct & Semi-ar diffusion & 77.48 & 1.0x & 30.61 & 1.0$\times$ & 51.40 & 1.0$\times$ & 22.32 & 1.0$\times$ & 38.71 & 1.0$\times$ & 39.20 & 1.0$\times$ \\ 
\midrule
LLaDA 8B Instruct & dLLM-cache & 78.54 & 4.0$\times$ & 29.01 & 4.6$\times$ & 51.98 & 3.7$\times$ & 22.52 & 4.7$\times$ & 37.19 & 2.5$\times$ & 39.60 & 4.0$\times$ \\ 
\midrule
LLaDA-8B-Instruct &  Fast dLLM & 78.50 & 13.8$\times$ & 29.73 & 9.4$\times$ & 50.74 & 16$\times$ & 33.20 & 6.6$\times$ & 43.30 & 7.0$\times$ & 28.20 & 12.3$\times$ \\ 

\midrule
dLLM-Var-pd & Pure diffusion & 77.81 & 1.0$\times$ & 32.92 & 1.0$\times$ & 53.71 & 1.0$\times$ & 32.58 & 1.0$\times$ & 39.94 & 1.0$\times$ & 41.8 & 1.0$\times$ \\ 
\midrule
\multirow{2}{*}{dLLM-Var-bd (ours)} & \multirow{2}{*}{Block diffusion} & 77.40 & \textbf{30.1$\times$} & \textbf{33.03} & \textbf{29.4$\times$} & 52.55 & \textbf{23.7$\times$} & 31.94 & \textbf{21.0$\times$} & \textbf{40.24} & \textbf{22.6$\times$} & \textbf{42.0} & \textbf{23.8$\times$} \\ 
& &  \textcolor{gray}{(-0.41)} &          \textcolor{LightGreen}{(+29.1$\times$)} &  \textcolor{LightGreen}{(+0.11)} &  \textcolor{LightGreen}{(+28.4$\times$)} &  \textcolor{gray}{(-1.16)} &  \textcolor{LightGreen}{(+22.7$\times$)} &  \textcolor{gray}{(-0.64)} &  \textcolor{LightGreen}{(+20.0$\times$)} &  \textcolor{LightGreen}{(+0.30)} &  \textcolor{LightGreen}{(+21.6$\times$)} &  \textcolor{LightGreen}{(+0.2)} &  \textcolor{LightGreen}{(+22.8$\times$)} \\
\bottomrule
\end{tabular}
}
\end{table}
% Please ensure your document preamble includes the following packages:
% \usepackage{multirow}
% \usepackage{booktabs}
% Please ensure your document preamble includes the following packages:
% \usepackage{multirow}
% \usepackage{booktabs}

% Please ensure your document preamble includes the following packages:
% \usepackage{multirow}
% \usepackage{booktabs}

% Please ensure your document preamble includes the following packages:
% \usepackage{multirow}
% \usepackage{booktabs}
\begin{table}[tb!]
\centering
% --- Table caption and label ---
\caption{Accuracy and speed-up comparison for dLLM-Var with different methods and lengths on the GSM8K and MBPP tasks. For Pure diffusion, the length represents the fixed generation quota.}
\label{tab:diffusion_comparison_final}
\resizebox{0.8\textwidth}{!}{
\begin{tabular}{lccccc}
\toprule
% --- Header row 1: Defines Inference Method, Length, and Tasks ---
\multirow{2}{*}{\textbf{Inference Method}} & \multirow{2}{*}{\textbf{Generation Length}} & \multicolumn{2}{c}{\textbf{GSM8K (5-shot)}} & \multicolumn{2}{c}{\textbf{MBPP (3-shot)}} \\
\cmidrule(lr){3-4} \cmidrule(lr){5-6}
% --- Header row 2: Defines Accuracy and Speed Up for each task ---
& & \textbf{Accuracy} & \textbf{Speed} & \textbf{Accuracy} & \textbf{Speed} \\
\midrule
% --- Data for Pure Diffusion ---
\multirow{5}{*}{Pure diffusion}
& 64 & 12.35 & 21.4$\times$ & 38.40 & 19.8$\times$ \\
& 128 & 51.93 & 10.1$\times$ & 42.00 & 9.1$\times$ \\
& 256 & 74.98 & 4.8$\times$ & 42.00 & 4.4$\times$ \\
& 512 & 75.96 & 2.3$\times$ & 43.20 & 2.1$\times$ \\
& 1024 & 77.81 & 1.0$\times$ & 41.80 & 1.0$\times$ \\
\midrule
% --- Data for Block Diffusion ---
Block diffusion & - & 77.40 & \textbf{30.1$\times$} & 42.00 & \textbf{23.8$\times$} \\
\bottomrule
\end{tabular}
}
\end{table}

\subsection{Experiments Result}

To validate the effectiveness of our proposed dLLM-Var, we conduct a comprehensive evaluation of its performance against several leading dLLM methodologies, including the baseline LLaDA-8B-Instruct, dLLM-Cache, and Fast dLLM. Our experiments focus on two critical metrics: inference speed and generation accuracy, with the results presented in Figure~\ref{fig:acc_speed}.

As illustrated in Figure~\ref{fig:acc_speed} (a), dLLM-Var achieves a dramatic improvement in inference efficiency. It demonstrates a remarkable speed-up, peaking at 30.1× relative to the standard LLaDA-8B-Instruct model. This performance significantly outpaces other advanced methods such as Fast dLLM and dLLM-Cache, confirming that our approach effectively leverages parallelism and KV caching without the overhead of complex attention mechanisms.

Crucially, these substantial gains in speed do not compromise the model's generative quality. Figure~\ref{fig:acc_speed} (b) shows the accuracy comparison across six diverse benchmarks: GSM8K, GPQA, BBH, MATH, HumanEval, and MBPP. dLLM-Var consistently delivers competitive or superior performance. For instance, it outperforms all baseline models on challenging benchmarks like BBH (52.55), GPQA (33.03), and MBPP (42.0). While its performance is on par with other methods in reasoning tasks like GSM8K, its robust results across different domains underscore that our training and inference paradigm successfully preserves, and in many cases enhances, the model's core capabilities.

\begin{figure}[tb!]
    \centering
    \includegraphics[width=0.99\linewidth]{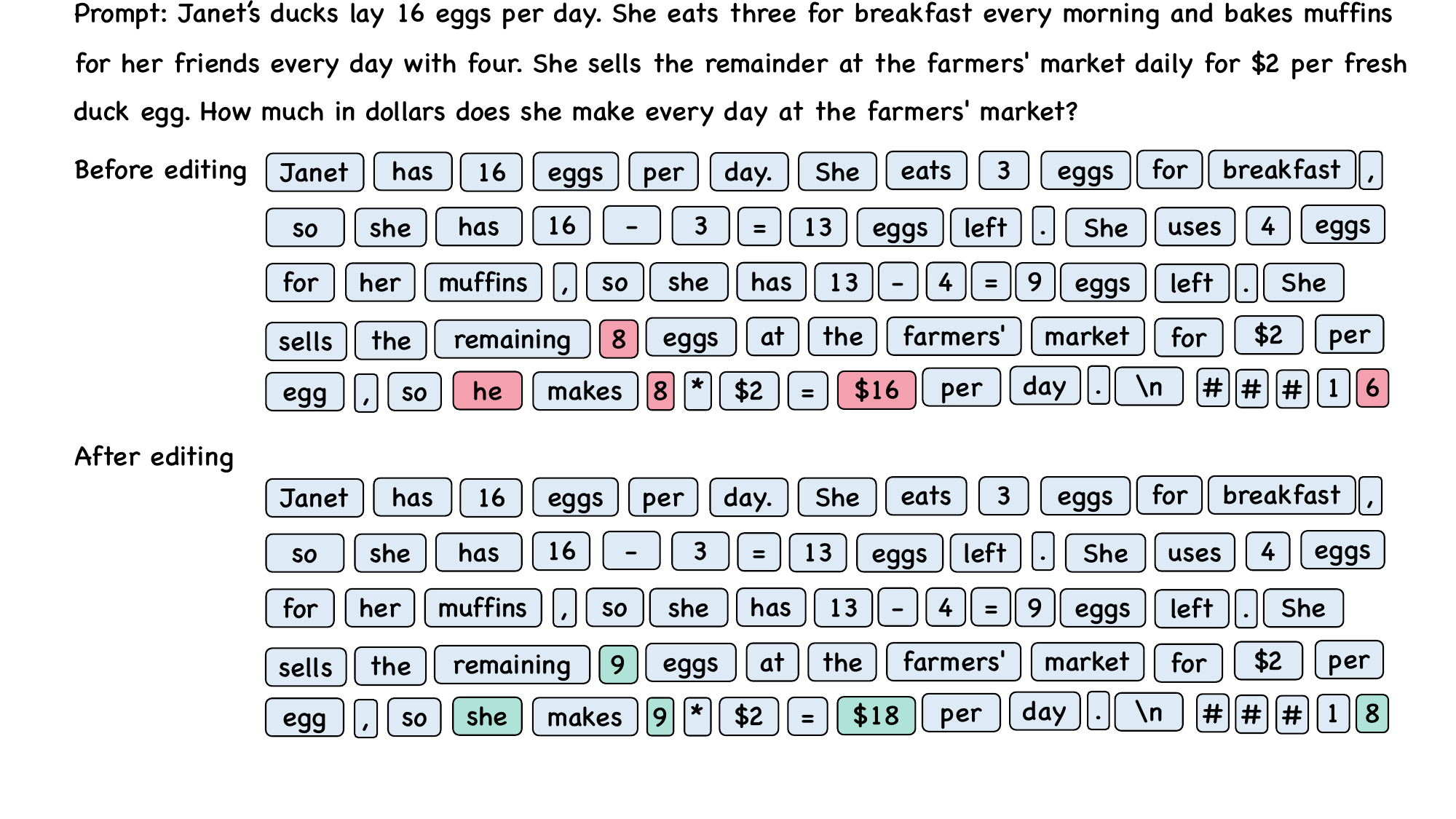}
    \caption{Demonstration of dLLM-Var's Self-Correction Capability. It illustrates the model's ability to refine its initial output. In the "Before editing" phase, the model makes both calculation error and  grammatical error. In the "After editing" phase, the model corrects these mistakes, adjusting the calculation to the correct 9 eggs and changing the pronoun to "she", resulting in the correct final answer of 18.}
    \label{fig:editing}
    
\end{figure}

\noindent\textbf{The Ability of Editing}: Our experiments also highlight the self-correction capabilities of dLLM-Var. As shown in Figure ~\ref{fig:editing}, the model can identify and rectify errors within its own generations. When tasked with a reasoning problem, an initial version of the model's response contained both logical and grammatical inaccuracies. Through its iterative refinement process, dLLM-Var revisited and amended this output to produce the correct and coherent final answer. This ability to self-correction demonstrates a promising capacity for reasoning about and improving its own output, marking a crucial step towards more reliable and accurate generative models.

\section{Conclusion and Future Works}
dLLM-Var revolutionizes diffusion-based LLMs by shattering fixed-length barriers and unlocking seamless KV cache reuse through an elegant training method: fixed EOS masking and multi-sample packing. With this work, open-source dLLMs transcend academic novelty, unlocking real-world industrial viability. While dLLM-Var marks a significant step towards making dLLMs practical, several exciting avenues for future research remain. Our current approach treats previously generated blocks as fixed, but this unidirectional process forgoes the opportunity for self-correction based on newly generated context. A promising direction is to explore mechanisms for iterative refinement and editing of generated content. As the model gains a richer contextual understanding from subsequent blocks, it could re-evaluate and selectively re-generate portions of earlier output to correct inaccuracies and improve coherence. The full attention mechanism employed in dLLM-Var is a key enabler for this research, allowing unrestricted access to all tokens. Future work could focus on developing efficient algorithms to decide "when" and "what" to edit, balancing the trade-off between quality gains and computational cost. Such advancements could unlock a new level of performance and reliability for dLLMs, further closing the gap with their autoregressive counterparts.

% 参考文献部分
% \nocite{*}  % 强制显示所有 .bib 文件中的引用
\bibliographystyle{unsrt}  
\bibliography{references}

\end{document}